\documentclass{article}

\PassOptionsToPackage{numbers, compress}{natbib}

    \usepackage[preprint]{neurips_2025}

\usepackage[utf8]{inputenc} 
\usepackage[T1]{fontenc}    
\usepackage{url}            
\usepackage{booktabs}       
\usepackage{amsfonts}       
\usepackage{nicefrac}       
\usepackage{microtype}      
\usepackage[svgnames]{xcolor}         
\usepackage{graphicx}
\usepackage{multirow}
\usepackage{amsmath}
\usepackage{wrapfig}
\usepackage{pifont}
\definecolor{redlinkcolor}{rgb}{0.79607843, 0.25098039, 0.25882353}
\definecolor{bluecitecolor}{rgb}{0,0.36,0.69}
\usepackage[colorlinks=true,linkcolor=redlinkcolor,citecolor=bluecitecolor,urlcolor=bluecitecolor]{hyperref} 

\title{PAID: Pairwise Angular-Invariant Decomposition for Continual Test-Time Adaptation}

\author{%
Kunyu Wang, Xueyang Fu, Yuanfei Bao, 
Chengjie Ge, Chengzhi Cao, \\ \textbf{Wei Zhai}, \textbf{Zheng-Jun Zha}\thanks{Corresponding Author} \\
University of Science and Technology of China\\
\texttt{kunyuwang@mail.ustc.edu.cn, zhazj@ustc.edu.cn}
}

\begin{document}

\maketitle

\begin{abstract}
\label{abstract}
  Continual Test-Time Adaptation (CTTA) aims to online adapt a pre-trained model to changing environments during inference. Most existing methods focus on exploiting target data, while overlooking another crucial source of information, the pre-trained weights, which encode underutilized domain-invariant priors. This paper takes the geometric attributes of pre-trained weights as a starting point, systematically analyzing three key components: magnitude, absolute angle, and pairwise angular structure. We find that the pairwise angular structure remains stable across diverse corrupted domains and encodes domain-invariant semantic information, suggesting it should be preserved during adaptation. Based on this insight, we propose \textbf{PAID} (\underline{\textbf{P}}airwise \underline{\textbf{A}}ngular-\underline{\textbf{I}}nvariant \underline{\textbf{D}}ecomposition), a prior-driven CTTA method that decomposes weight into magnitude and direction, and introduces a learnable orthogonal matrix via Householder reflections to globally rotate direction while preserving the pairwise angular structure. During adaptation, only the magnitudes and the orthogonal matrices are updated. PAID achieves consistent improvements over recent SOTA methods on four widely used CTTA benchmarks, demonstrating that preserving pairwise angular structure offers a simple yet effective principle for CTTA.
\end{abstract}

\section{Introduction}
\label{introduction}
Deep Neural Networks (DNNs) have achieved remarkable success in various computer vision tasks~\cite{mauricio2023comparing,zou2023object,minaee2021image}. However, when there exist domain discrepancies between training and testing environments~\cite{recht2019imagenet,hendrycks2019benchmarking,chen2019progressive,chen2020harmonizing,chen2021i3net}, directly applying a source pre-trained model may cause significant performance degradation, particularly when the target distribution is unpredictable and continually changing over time. This challenge in real-world scenarios has motivated the emergence of Continual Test-Time Adaptation (CTTA)~\cite{wang2022continual}, which aims to adapt source pre-trained models during inference to evolving test data, making it especially suitable for practical applications.

In CTTA, models rely on two key sources of information: the prior knowledge encoded in pre-trained source weights and the streaming data from the target domains. 
Most existing methods~\cite{wang2024search,liang2025comprehensive,xiao2024beyond} focus on adapting to the target data, while treating pre-trained weights as static initialization, with their potential largely overlooked.
However, these weights, learned from large-scale supervised training~\cite{deng2009imagenet,ridnik2021imagenet}, may encode transferable priors that remain invariant across domains. 
We posit that leveraging such invariances in parameter space can help address core challenges in CTTA, including catastrophic forgetting and error accumulation. 
This work explores this direction by uncovering domain-invariant components in pre-trained weights, offering a new perspective on knowledge retention and transfer for continual adaptation.

\subsection{Motivation}
\label{motivation}
Recent studies in hyperspherical learning~\cite{liu2017deep,liu2021orthogonal,liu2018learning,liu2018decoupled,tan2022hyperspherical} reveal that the angular component of neuron weights, rather than their magnitudes, primarily encodes discriminative semantic information crucial for visual recognition. Similar observations have been made in generative models~\cite{qiu2023controlling}, where the pairwise angular structure among neurons effectively preserve semantic consistency after fine-tuning on different domains. 
These findings suggest that beyond treating neural weights as individual scalars, the geometry of weight space, particularly the angular components, may encode invariant semantic priors derived from pre-training. Motivated by this, we ask: can the angular configuration of pre-trained weights serve as a stable semantic anchor during continual adaptation? To investigate this hypothesis, we design three sets of targeted experiments that isolate and examine the respective roles of magnitude and angular components in the adaptation process.
\footnote{More experimental details can be found in Appendix \ref{motivation_experiment_detail}}

Specifically, neuron weights, as exemplified by the linear projection matrix $W = [w_1, w_2, \dots, w_k] \in \mathbb{R}^{d \times k}$, can be decomposed into a magnitude $M \in \mathbb{R}^{1 \times k}$ and a unit-length direction $\hat{W} \in \mathbb{R}^{d \times k}$ :
\begin{equation}
\label{eq1}
    W = M \odot \hat{W}, \quad \text{where} \quad w_i = \|w_i\| \cdot \hat{w}_i, \quad \hat{w}_i = \frac{w_i}{\|w_i\|}, \quad \|\hat{w}_i\| = 1.
\end{equation}
Building on this view, we identify three geometric attributes of the weight space that are relevant to adaptation:
(1) the magnitude, which determines the scaling of feature responses;
(2) the absolute angle, referring to the orientation of each unit direction vector in the feature space, which may rotate during adaptation;
(3) the pairwise angular structure, defined as the set of angles between all pairs of unit direction vectors, capturing how weight vectors are arranged relative to one another.

\begin{figure}[t]
    \centering
    \includegraphics[width=1.0\linewidth]{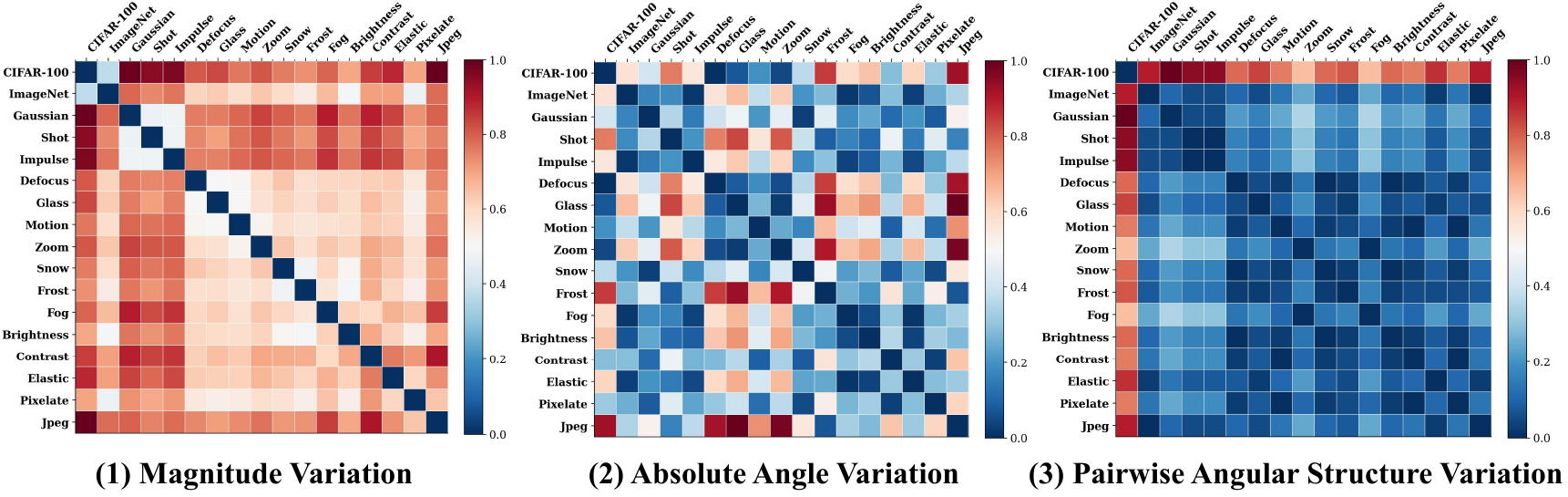}
    \caption{(\textbf{Experiment 1}) Visualization of cross-domain variation of three geometric properties. Pairwise angular structure remains stable under corruption but varies under semantic shift, suggesting it encodes semantic-relevant, domain-invariant information. In contrast, magnitude and absolute angle fluctuate irregularly across domains, reflecting domain-specific shifts.}
    \vspace{-0.4cm}
    \label{motivation1}
\end{figure}

\textbf{Experiment 1} examines the domain invariance of geometric attributes in weight space by analyzing their variation under different conditions. We use a ViT-Base model pre-trained on ImageNet and conduct two types of experiments: (1) Corruption, where the model performs test-time adaptation on each of the 15 corrupted domains in ImageNet-C; and (2) Semantic shift, where the model is fine-tuned on CIFAR-100 to induce semantic changes, serving as a reference. To quantify the average variation in geometric attributes, we define three metrics: (1) magnitude variation $\Delta \text{M}$, (2) absolute angle variation $\Delta \text{A}$, and (3) pairwise angular structure variation $\Delta \text{S}$, quantified using hyperspherical energy~\cite{liu2018learning,qiu2023controlling}, which refers to the sum of hyperspherical similarity between all pairwise neurons:
\begin{align}
    \Delta \text{M}(W^1, W^2) &= \frac{1}{k} \textstyle \sum_{n=1}^{k} \left| \left\| w_n^1 \right\| - \left\| w_n^2 \right\| \right|, \;
    \Delta \text{A}(W^1, W^2) = \frac{1}{k} \textstyle \sum_{n=1}^{k} \left( 1 - \cos(\hat{w}_n^{1}, \hat{w}_n^{2}) \right),
\end{align}
\begin{equation}
\Delta \text{S}(W^1, W^2) = \left| 
\textstyle \sum_{i \ne j} \left\| \hat{w}_i^1 - \hat{w}_j^1 \right\|^{-1}
-
\textstyle \sum_{i \ne j} \left\| \hat{w}_i^2 - \hat{w}_j^2 \right\|^{-1}
\right|, \quad \text{where } i,j \in \{1, \dots, k\},
\end{equation}

As shown in Fig. \ref{motivation1}, the pairwise angular structure remains consistently stable across all corruption domains, with minimal changes in hyperspherical energy. In contrast, it exhibits substantial variation under semantic shift. This contrast reveals that pairwise angular structure is well preserved under perturbations and varies in response to semantic changes. In contrast, both magnitude and absolute angle exhibit irregular variations across domains, indicating that their changes are driven by adaptation to domain-specific statistical differences rather than semantics. These findings provide strong empirical support that pairwise angular structure encodes semantic-relevant, domain-invariant properties, whereas magnitude and absolute direction are more responsive to domain-specific statistical shifts.

\begin{figure}[t]
    \centering
    \includegraphics[width=1.0\linewidth]{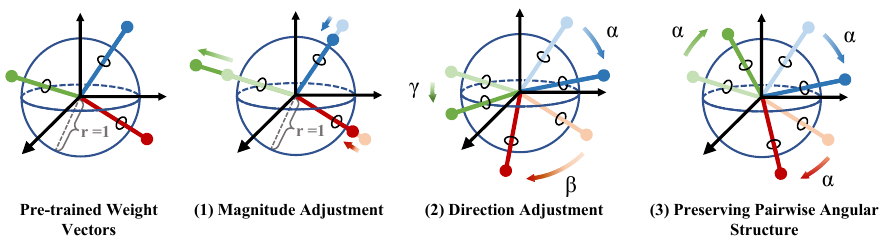}
    \caption{(\textbf{Experiment 2}) Illustration of three update strategies. (1) Scaling the magnitude of each vector; (2) Independently rotating each vector, altering their absolute angle; (3) Jointly rotating all vectors while preserving their pairwise angular structure.}
    \label{teasor}
    \vspace{-0.2cm}
\end{figure}
\begin{figure}[t]
    \centering
    \includegraphics[width=1.0\linewidth]{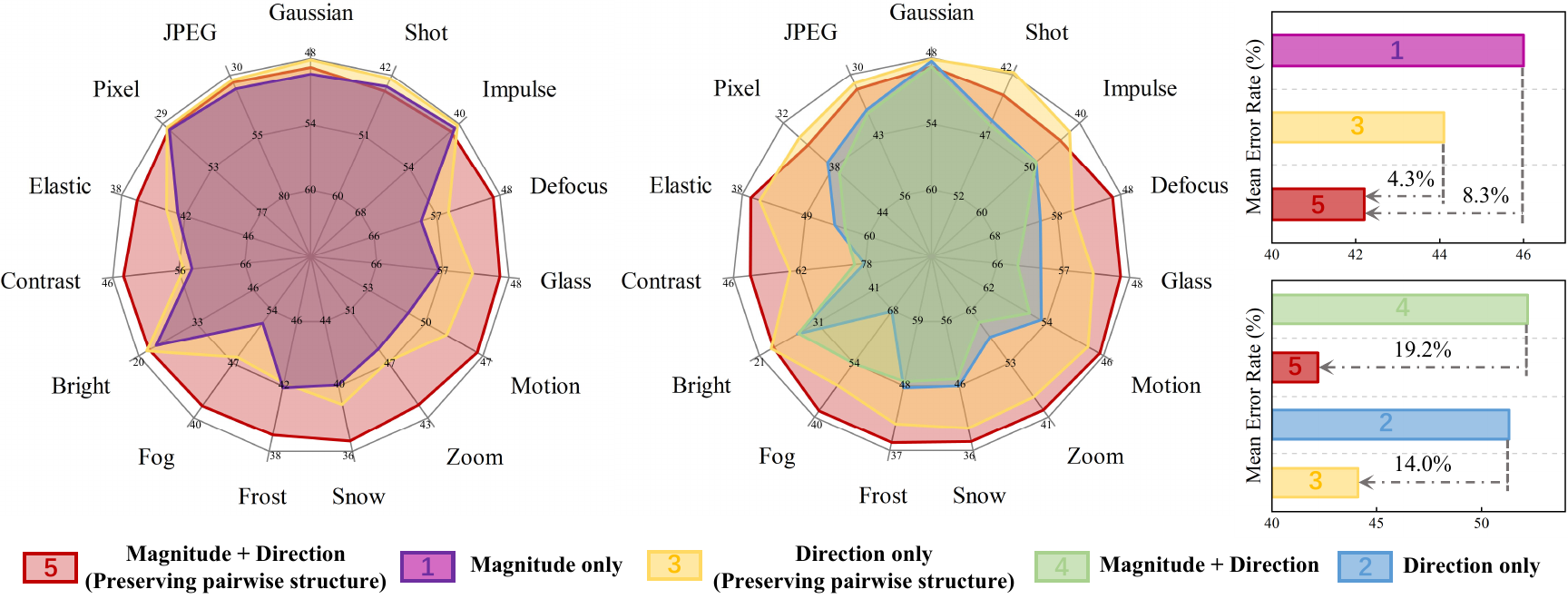}
    \caption{(\textbf{Experiment 2}) Radar and bar charts showing classification error rates across 15 corruption domains and their mean. Comparisons between settings (5 vs 1, 3; 5 vs 4; 3 vs 2) show that pairwise angular structure is a domain-invariant component worth preserving, while magnitude and direction, when constrained by fixed angular structure, are domain-specific and beneficial for adaptation.} 
    \vspace{-0.4cm}
    \label{motivation2}
\end{figure}

\textbf{Experiment 2} conducts an ablation study to clarify the contribution of geometric attributes to adaptation. Based on the decomposition, we design three update strategies, as shown in Fig. \ref{teasor}: (1) magnitude adjustment, enabling learning of the magnitude matrix $M$; (2) direction adjustment, enabling learning of unit direction matrix $\hat{W}$; and (3) direction adjustment while preserving pairwise angular structure, implemented via orthogonal rotation~\cite{yuan2024bridging,dong2024efficient,householder1958unitary} of direction matrix. Note that changing the pairwise angular structure inherently changes absolute angle, but not vice versa. Thus, (3) represents a constrained subset of (2). Using these strategies, we construct five settings for continually adapting a ViT-Base model (pre-trained on ImageNet) to the 15 ImageNet-C domains: (1) magnitude only; (2) direction only; (3) direction only while preserving pairwise structure; (4) magnitude + direction; and (5) magnitude + direction while preserving pairwise structure.

As shown in Fig. \ref{motivation2}, setting (5) outperforms settings (1) and (3), while settings (2) and (4), both of which alter the pairwise angular structure, lead to a clear performance drop compared to (3) and (5). This contrast underscores the role of pairwise angular structure as a domain-invariant component that should be preserved during adaptation. In contrast, magnitude and direction, when adjusted under the constraint of maintaining relative angular geometry, serve as domain-specific components that are adaptable and beneficial for enhancing performance in CTTA.

\begin{figure}[htbp]
    \centering
    \includegraphics[width=1.0\linewidth]{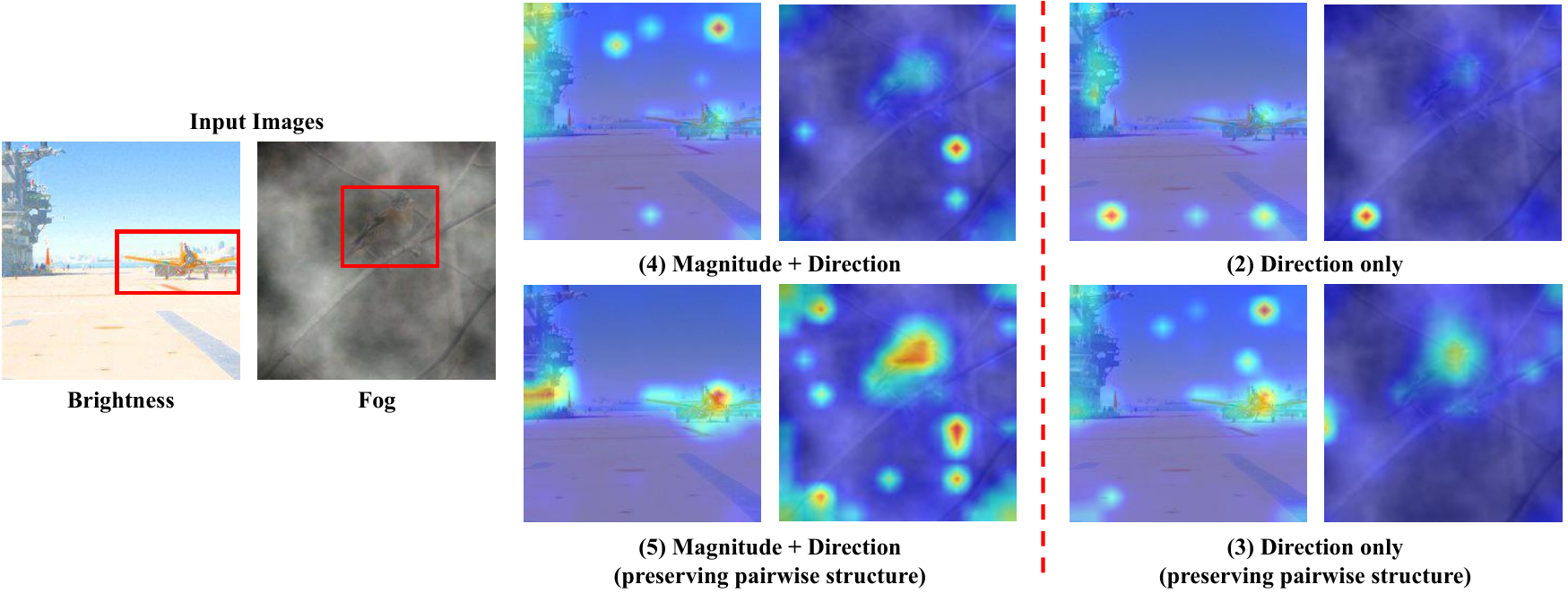}
    \caption{(\textbf{Experiment 3}) Attention map visualizations for two setting pairs (5 vs 4; 3 vs 2). The comparison further supports the semantic relevance and domain invariance of pairwise angular structure in CTTA.}
    \vspace{-0.4cm}
    \label{motivation3}
\end{figure}

\textbf{Experiment 3} further substantiates the earlier conclusions from a perceptual perspective by visualizing attention maps under the five settings in Experiment 2. As shown in Fig. \ref{motivation3}, preserving the pairwise angular structure allows the model to focus on crucial semantic regions. In contrast, settings (2) and (4) that alter this structure lead to diffused and misaligned attention, with the model failing to capture core object information. This comparison reinforces the critical role of preserving pairwise angular structure for achieving cross-domain invariance.

\subsection{Contribution}
These three sets of experiments—statistical analysis, functional validation, and visual interpretation—collectively reveal a key insight: the pairwise angular structure of neural weights encodes a semantically relevant and domain-invariant prior derived from pre-training, which should be preserved during CTTA. In contrast, the magnitude and absolute angle, when adjusted under the constraint of preserving the pairwise angular structure, serve as domain-specific components that enable effective adaptation to the target domains.

Building on this insight, we propose Pairwise Angular-Invariant Decomposition (PAID) for CTTA. PAID explicitly decomposes pre-trained weights into magnitude and direction matrices. To maintain the pairwise angular structure during adaptation, we introduce a learnable orthogonal matrix constructed via Householder transformations, enabling global rotation of directions without altering their relative angular configuration. Leveraging this orthogonality, we freeze the original direction matrix and update only the magnitude and the injected orthogonal matrix, ensuring structure-preserving adaptation.
In summary, the contributions can be summarized as follows:
\begin{itemize}
\item We identify the pairwise angular structure of pre-trained weights as a domain-invariant semantic prior that should be preserved during CTTA, supported by statistical, functional, and visual analyses.

\item We propose PAID, a novel prior-driven CTTA method that preserves the pairwise angular structure of weights while enabling controlled adaptation of their magnitudes and directions through orthogonal transformations.

\item PAID achieves consistent improvements over recent SOTA methods on four standard CTTA benchmarks, demonstrating strong effectiveness and generalizability.
\end{itemize}

\section{Related Work}

\noindent
{\bf Continual Test-time Adaptation (CTTA)}
aims to online adapt a source pre-trained model to handle a sequence of target domains. Existing learning paradigms can be broadly categorized into three types~\cite{wang2024search,liang2025comprehensive,xiao2024beyond}. Optimization-based methods aim to adjust pre-trained models by designing new objectives, including statistics calibration~\cite{yuan2023robust,gong2022note,tomar2024mixing}, consistency regularization~\cite{tomar2023tesla,song2023ecotta,liu2023vida,yang2024versatile}, entropy minimization~\cite{wang2020tent,niu2022efficient,zhao2023delta}, and pseudo-labeling~\cite{jang2022test,zeng2024rethinking,chen2022contrastive}. Data-based methods focus on enhancing data diversity or mitigating the impact of distributional shifts. Typical strategies include data augmentation~\cite{marsden2024universal,zhang2022memo,fleuret2021test} and memory bank~\cite{gong2023sotta,wang2023feature}. Model-based methods enhance adaptation by modifying or extending the model architecture, including module addition~\cite{jang2022test,liu2023vida}, module substitution~\cite{iwasawa2021test}, and prompt-based mechanisms~\cite{shu2022test,gan2023decorate,gao2022visual,zhang2024dynamic}. While similar in form to model-based methods, our method introduces a novel perspective by leveraging transferable priors encoded in pre-trained weights. We exploit the pairwise angular structure as a domain-invariant prior to enhance CTTA.

\noindent
{\bf Parameter-Efficient Fine-Tuning (PEFT)}~\cite{han2024parameter} reduces adaptation cost by freezing most large model parameters and updating only a small, task-specific subset. Representative methods include adapter-based methods~\cite{houlsby2019parameter}, low-rank adaptation~\cite{hu2022lora,zhang2023adalora,dong2023efficient} and prompt tuning~\cite{qin2021learning,jia2022visual}. Among them, DoRA~\cite{liu2024dora} and OFT~\cite{qiu2023controlling} are particularly insightful: DoRA decomposes weights into magnitude and direction, applying low-rank updates solely to the direction for near full-tuning expressiveness; OFT adopts layer-shared orthogonal transformations to fine-tune only the direction of weights, preserving the angular structure critical for semantic consistency in generative models. Inspired by these, we introduce the idea of weight decomposition into the field of test-time adaptation, rethinking it from the lens of generalization. Through a systematic analysis of pre-trained weight space, we find that the pairwise angular structure remains stable across domains and encodes domain-invariant semantics. This observation, supported by statistical, functional, and visual evidence, motivates us to explicitly preserve angular structure during CTTA.

\section{Methodology}
\subsection{Preliminaries}

\begin{wrapfigure}{r}{0.5\textwidth}
    \centering
    \includegraphics[width=\linewidth]{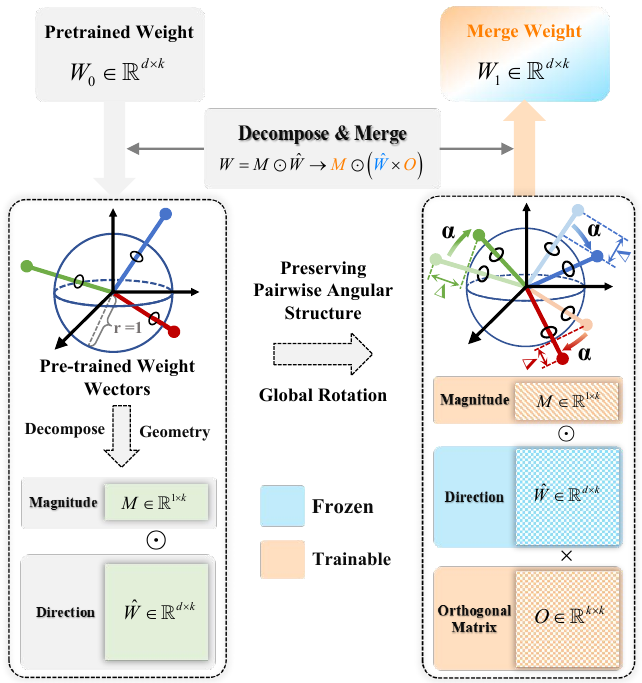}
    \caption{PAID decomposes pre-trained weights into magnitude and direction. To preserve pairwise angular structure, we introduce a learnable orthogonal matrix, enabling global rotation. Only magnitude and orthogonal matrices are updated during adaptation.}
    \label{fig:main}
    \vspace{-0.2cm}
\end{wrapfigure}

Given a model \( f_\theta \) pre-trained on the source domain \( D_s = \{x_s, y_s\} \), our goal is to adapt this model to a sequence of continually changing target domains \( \{D_t^1, D_t^2, \dots, D_t^N\} \). In an online setting, the model \( f_\theta \) processes a sequence of test data batches \( \{B_t\}_{t=1}^\infty \), with each batch \( B_t \) arriving at time step \( t \). Consistent with prior work~\cite{wang2022continual}, we assume that all samples in a batch \( B_t \) come from the same target domain, though the domain identity is unknown. At each time step \( t \), CTTA aims to adapt the model parameters from \( \theta_t \) to \( \theta_{t+1} \) by learning from the current batch \( B_t \), thereby enhancing performance on subsequent batches.

A Vision Transformer model consists of multiple encoder layers, each containing a Multi-Head Attention (MHA) block and a Feed-Forward Network (FFN) block. The MHA block utilizes three key linear weights \( W_q \) for the query, \( W_k \) for the key, and \( W_v \) for the value to compute attention scores and aggregate weighted values from normalized feature representations. In addition, a linear weight \( W_o \) combines the outputs of all attention heads. In the FFN block, the input is processed through two linear weights \( W_{m_1} \) and \( W_{m_2} \) with a GELU function applied between them. 
This work focuses on decomposing the pre-trained weights of these linear layers \( W_q \), \( W_k \), \( W_v \), \( W_o \), \( W_{m_1} \) and \( W_{m_2} \), preserving their domain-invariant priors to enhance CTTA task. Fig.~\ref{fig:main} provides an overview of our method.

\subsection{Pairwise Angular-Invariant Decomposition (PAID)}
Following Eq.~\ref{eq1}, a linear projection matrix \( W = [w_1, w_2, \dots, w_k] \in \mathbb{R}^{d \times k} \) can be decomposed into a magnitude matrix \( M \in \mathbb{R}^{1 \times k} \) and a direction matrix \( \hat{W} = [\hat{w}_1, \hat{w}_2, \dots,\hat{w}_k] \in \mathbb{R}^{d \times k} \) as:
\begin{equation}
    W = M \odot \hat{W}, \quad \text{where} \quad w_i = \|w_i\| \cdot \hat{w}_i, \quad \|\hat{w}_i\| = 1.
\end{equation}
Empirical evidence suggests that the pairwise angular structure of pre-trained weights, defined as the angles between all pairs of unit direction vectors, encodes a semantically meaningful and domain-invariant prior. This structure should be preserved during the CTTA process. In contrast, magnitude and absolute angle, when adapted under the constraint of preserving this structure, serve as domain-specific components that support effective adaptation. Accordingly, we make the magnitude matrix $M$ learnable while freezing the original direction matrix $\hat{W}$. To enable directional adaptation without disrupting angular structure, we introduce a learnable orthogonal matrix $O \in \mathbb{R}^{k \times k}$, which performs global rotations while preserving pairwise angular structure:
\begin{equation}
    M \odot \hat{W}
    \longrightarrow
    \textcolor{Orange}{M} \odot \left( \textcolor{cyan!70!white}{\hat{W}} \cdot \textcolor{Orange}{O} \right),
\end{equation}
A real square matrix \(O \in \mathbb{R}^{k \times k}\) is orthogonal if it satisfies
\(O^{\top} O = I\).
Such matrix represent distance‑preserving linear transformation,
including rotation and reflection.
For any vector \(x \in \mathbb{R}^{k}\), the transformation \(x \mapsto O x\)
preserves the Euclidean norm, that is, \(\lVert O x \rVert = \lVert x \rVert\).
For any pair of vectors $(x, y)$, it also preserves inner products:
\(
\langle O x,\; O y \rangle \;=\; \langle x,\; y \rangle .
\)
As a result, the relative geometry among the vectors remains unchanged. Therefore, applying an orthogonal transformation to a set of vectors performs a global rotation without altering their pairwise angular structure.

To construct the orthogonal transformation, we adopt the Householder reflection formulation~\cite{householder1958unitary,yuan2024bridging,dong2024efficient}.  
A Householder reflection is a linear transformation that reflects a vector across a hyperplane perpendicular to a unit vector.  
Given \( u \in \mathbb{R}^{k} \) with \( \|u\|_2 = 1 \), the corresponding Householder matrix is defined as:
\begin{equation}
    H = I - 2uu^\top,
\end{equation}
where \( H \in \mathbb{R}^{k \times k} \) is orthogonal and symmetric, satisfying \( H^\top H = I \) and \( \det(H) = -1 \), indicating a reflection.  
The transformation preserves Euclidean norms and inner products, i.e.,
\begin{equation}
    \|Hx\|_2 = \|x\|_2, \quad \langle Hx, Hy \rangle = \langle x, y \rangle, \quad \forall x, y \in \mathbb{R}^k,
\end{equation}
thereby maintaining both vector lengths and pairwise angles.
Since Householder matrices are orthogonal, and the product of orthogonal matrices remains orthogonal, a general orthogonal matrix \( O \in \mathbb{R}^{k \times k} \) can be constructed as a chain of \( r \) Householder reflections:
\begin{equation}
    O = \prod_{i=1}^{r} H_i = \prod_{i=1}^{r} (I - 2 u_i u_i^\top), \quad u_i \in \mathbb{S}^{k-1}.
\end{equation}
This parameterization is expressive: when \( r = k \), it can represent any element in the orthogonal group \( \mathrm{O}(k) \), while smaller \( r \) yields a trade-off between representational capacity and efficiency. Therefore, we adopt such a chain to construct the learnable orthogonal transformation used for structure-preserving adaptation and $r$ denotes orthogonal matrix coefficient.

\subsection{Optimization Objective}
Following previous CTTA work~\cite{mirza2023actmad,niu2024test,zhang2025ot}, we update PAID by aligning feature distributions between the source and target domains. Specifically, we pre-compute the source feature statistics, including the mean \( \mu_s \) and standard deviation \( \sigma_s \), from a small source subset \( B_s \subset D_s \). Let \( Z_t^T \) represent the extracted features of the current target batch \( B_t^T \), with corresponding statistics \( \mu_t^T \) and \( \sigma_t^T \).
Our goal is to minimize the discrepancy between the source and target domain statistics by aligning their means and standard deviations. We formalize this alignment with the following loss function:
\begin{equation}
    \mathcal{L} = || \mu_s - \mu_t^T ||_2 + \lambda || \sigma_s - \sigma_t^T ||_2,
\end{equation}
where $\lambda$ denotes the loss balancing coefficient. Note that the source data is only accessed prior to CTTA, with merely 500 images enough to ensure performance. No further access to the source domain is required during the CTTA process. Additionally, the storage requirements for these statistics are minimal and negligible.

\subsection{Intuitive Explanation of PAID}
To further explain the intuition for why PAID works, we draw an analogy from the frequency domain decomposition of images. In the Fourier transform~\cite{nussbaumer1982fast}, an image can be decomposed into a magnitude and a phase spectrum.  The magnitude captures energy distribution, contrast, and intensity, and is highly sensitive to domain-specific variations such as noise and style. In contrast, the phase encodes angular alignment with Fourier bases, which determines the structural layout and semantic content.
The separation between semantic structure and domain-specific appearance has been widely explored in frequency-based domain generalization~\cite{yang2020fda,jeon2021feature,lee2023decompose,huang2021fsdr,lin2023deep}. For instance, APR~\cite{chen2021amplitude} reveals that networks are more sensitive to magnitude perturbations, while phase is crucial for retaining semantic information and achieving robust recognition. FACT~\cite{xu2021fourier} verifies the domain-invariant property of phase by showing that transferring phase yields better generalization than amplitude.
This perspective supports our design in PAID. The pairwise angular structure, akin to the phase spectrum, captures semantics and remains invariant across domains. PAID preserves the angular structure while adapting the magnitude and direction via constrained orthogonal transformations, thus achieving structure-preserving adaptation across domains.

\section{Experiments}
\label{experiments}

In Section~\ref{main_results}, we compare our method with state-of-the-art approaches on both classification and segmentation CTTA benchmarks. Section~\ref{ablation_studies} analyzes the effectiveness of our design choices and investigates the influence of the injected matrix coefficient, test batch size, number of source examples, and computational overhead. Additional ablations, including the effect of the loss coefficient, the choice of injection layer, and a 10-round classification CTTA evaluation, are provided in Appendix~\ref{additional_ablation_studies}.

\begin{table}[t]
  \centering
  \setlength{\tabcolsep}{4pt}
  \caption{Mean classification error rate (\%) and gain (\%) on ImageNet-to-ImageNet-C using ViT-Base. \textbf{Bold} indicates the best performance. Fine-grained performances are shown in Appendix \ref{fine-grained}.}
  \resizebox{\linewidth}{!}{
    \begin{tabular}{l|ccccccccccc}
      \toprule
      Method & Source & Pseudo~\cite{lee2013pseudo} & TENT~\cite{wang2020tent} & CoTTA~\cite{wang2022continual} & VDP~\cite{gan2023decorate} & SAR~\cite{niu2023towards} & RoTTA~\cite{yuan2023robust} & EcoTTA~\cite{song2023ecotta} & ViDA~\cite{liu2023vida} & C-MAE~\cite{liu2024continual} & Ours \\
      \midrule
      Mean$\downarrow$ & 55.8 & 66.8 & 51.0 & 54.8 & 50.0 & 45.6 & 48.2 & 48.0 & 43.4 & 42.5 & \textbf{42.2} \\
      Gain$\uparrow$ & 0.0 & -11.0 & +4.8 & +1.0 & +5.8 & +10.2 & +7.6 & +7.8 & +12.4 & +13.3 & \textbf{+13.6} \\
      \bottomrule
    \end{tabular}
  }
  \label{tab:imagenet-mean-gain}
\end{table}

\begin{table}[t]
  \centering
  \setlength{\tabcolsep}{5pt}
  \caption{Mean classification error rate (\%) and gain (\%) on CIFAR100-to-CIFAR100-C using ViT-Base. \textbf{Bold} indicates the best performance. Fine-grained performances are shown in Appendix \ref{fine-grained}.}
  \resizebox{\linewidth}{!}{
    \begin{tabular}{l|cccccccc}
      \toprule
      Method & Source & Pseudo~\cite{lee2013pseudo} & TENT~\cite{wang2020tent} & CoTTA~\cite{wang2022continual} & VDP~\cite{gan2023decorate} & ViDA~\cite{liu2023vida} & C-MAE~\cite{liu2024continual} & Ours \\
      \midrule
      Mean$\downarrow$ & 35.4 & 33.2 & 32.1 & 34.8 & 32.0 & 27.3 & 26.4 & \textbf{24.9} \\
      Gain$\uparrow$ & 0.0 & +2.2 & +3.3 & +0.6 & +3.4 & +8.1 & +9.0 & \textbf{+10.5} \\
      \bottomrule
    \end{tabular}
  }
  \vspace{-0.4cm}
  \label{tab:cifar100-mean-gain}
\end{table}

\subsection{Experimental Setup}

\noindent
{\bf Datasets.}
We evaluate our method on three classification CTTA benchmarks: CIFAR10-to-CIFAR10C, CIFAR100-to-CIFAR100C~\cite{krizhevsky2009learning}, and ImageNet-to-ImageNet-C~\cite{hendrycks2019benchmarking}. 
In the classification tasks, we follow the sequential adaptation process described in~\cite{song2023ecotta}, where the pre-trained source model adapts to each of the 15 target domains, each defined by the highest corruption severity. Online prediction results are immediately assessed after processing the input.
For segmentation CTTA, we assess our method on Cityscapes-to-ACDC, where Cityscapes~\cite{cordts2016cityscapes} serves as the source domain and ACDC~\cite{sakaridis2021acdc} as the target domains, which includes images captured under four distinct unobserved visual conditions: Fog, Night, Rain, and Snow.
To simulate continual environmental changes, we cyclically iterate through the same sequence of target domains (Fog $\rightarrow$ Night $\rightarrow$ Rain $\rightarrow$ Snow) three rounds, reflecting real-world scenarios.

\noindent
{\bf Methods Compared.}
We compare our method against several strong CTTA baselines, including Source, Pesudo~\cite{lee2013pseudo}, TENT~\cite{wang2020tent}, CoTTA~\cite{wang2022continual}, DePT~\cite{gao2022visual}, VDP~\cite{gan2023decorate}, SAR~\cite{niu2023towards}, RoTTA~\cite{yuan2023robust}, EcoTTA~\cite{song2023ecotta}, ViDA~\cite{liu2023vida}, and C-MAE~\cite{liu2024continual}. "Source" represents the use of the pre-trained model for adaptation without any specific method. The selection of comparison methods is based on their open-source availability and representativeness, with implementations and results drawn from publicly available codebases, paper descriptions, and established benchmarks.

\noindent
{\bf Implementation Details.}
For the classification CTTA tasks, we use ViT-base~\cite{dosovitskiy2020image} as the backbone model, resizing input images to 384×384 for CIFAR10-C and CIFAR100-C, and to 224×224 for the ImageNet-C benchmark. For the segmentation CTTA task, we employ Segformer-B5~\cite{xie2021segformer} pre-trained on the Cityscapes dataset as the source model, down-sampling input images from 1920$\times$1080 to 960$\times$540. The AdamW~\cite{loshchilov2017decoupled} optimizer is used with parameters ($\beta_1$, $\beta_2$) = (0.9, 0.999). Hyper-parameters for CIFAR10-C, CIFAR100-C, ImageNet-C, and ACDC are set as follows: learning rate \{8e-5, 8e-5, 2e-4, 1e-4\}, batch size \{64, 64, 64, 1\}, orthogonal matrix coefficient \{12, 12, 12, 12\}, and loss coefficient \{1.0, 1.0, 0.1, 1.0\}. To initialize the learnable parameters, we perform warm-up iterations on classification datasets. For linear layer selection, we inject all linear layers, including those in the multi-head attention block (q, k, v, o) and the MLP block (m). The number of source examples is set to 500. The experiments are conducted on the NVIDIA RTX 3090 GPU. All ablation studies are conducted on ImageNet-to-ImageNet-C unless otherwise specified.

\begin{table}[t]
  \centering
  \setlength{\tabcolsep}{5pt}
  \caption{Mean classification error rate (\%) and gain (\%) on CIFAR10-to-CIFAR10-C using ViT-Base. \textbf{Bold} indicates the best performance. Fine-grained performances are shown in Appendix \ref{fine-grained}.}
  \resizebox{\linewidth}{!}{
    \begin{tabular}{l|cccccccc}
      \toprule
      Method & Source & Pseudo~\cite{lee2013pseudo} & TENT~\cite{wang2020tent} & CoTTA~\cite{wang2022continual} & VDP~\cite{gan2023decorate} & ViDA~\cite{liu2023vida} & C-MAE~\cite{liu2024continual} & Ours \\
      \midrule
      Mean$\downarrow$ & 28.1 & 26.9 & 23.5 & 24.6 & 24.1 & 20.7 & 12.6 & \textbf{11.0} \\
      Gain$\uparrow$ & 0.0 & +1.2 & +4.6 & +3.5 & +4.0 & +7.4 & +15.5 & \textbf{+17.1} \\
      \bottomrule
    \end{tabular}
  }
  \label{tab:cifar10-mean-gain}
\end{table}

\begin{table}[t]
  \centering
  \caption{Mean mIoU (\%) and gain (\%) on Cityscapes-to-ACDC (3-round average) using Segformer-B5. \textbf{Bold} indicates the best performance. Fine-grained performances are shown in Appendix \ref{fine-grained}.}
  \setlength{\tabcolsep}{4pt}
  \resizebox{\linewidth}{!}{
    \begin{tabular}{l|cccccccccc}
      \toprule
      Metric & Source & TENT~\cite{wang2020tent} & CoTTA~\cite{wang2022continual} & DePT~\cite{gao2022visual} & VDP~\cite{gan2023decorate} & SAR~\cite{niu2023towards} & EcoTTA~\cite{song2023ecotta} & ViDA~\cite{liu2023vida} & C-MAE~\cite{liu2024continual} & Ours \\
      \midrule
      Mean$\uparrow$ & 56.7 & 55.7 & 58.6 & 53.4 & 58.2 & 57.0 & 55.8 & 61.9 & 61.8 & \textbf{62.2} \\
      Gain$\uparrow$ & 0.0 & -1.0 & +1.9 & -3.3 & +1.5 & +0.3 & -0.9 & +5.2 & +5.1 & \textbf{+5.5} \\
      \bottomrule
    \end{tabular}
  }
  \label{tab:acdc-mean-gain}
  \vspace{-0.4cm}
\end{table}

\subsection{Results on Benchmark Datasets}
\label{main_results}
Table~\ref{tab:imagenet-mean-gain} reports classification error rates on ImageNet-C under corruption severity level 5. Our method achieves the lowest mean error of 42.2\%, outperforming strong baselines such as SAR (45.6\%), ViDA (43.4\%), and C-MAE (42.5\%). Compared to the source model, our method yields a substantial improvement of +13.6\%. 
In Table~\ref{tab:cifar100-mean-gain}, we present results on CIFAR100-C. Our method achieves the best overall performance with a mean error of 24.9\%, surpassing recent leading methods including ViDA (27.3\%) and C-MAE (26.4\%). The performance gain over the source model reaches +10.5\%.
Table~\ref{tab:cifar10-mean-gain} shows the results on the simpler CIFAR10-C dataset. Our method achieves a mean error of 11.0\%, which is a +17.1\% improvement over the source model and also outperforms the best baseline (C-MAE at 12.6\%). This confirms the effectiveness of our adaptation mechanism even on small-scale datasets.
For the segmentation task, Table~\ref{tab:acdc-mean-gain} reports the average mIoU on the Cityscapes-to-ACDC benchmark over three consecutive adaptation rounds. Our method achieves the highest average mIoU of 62.2\%, showing robust performance across rounds and delivering a consistent +5.5\% improvement over the source model. 

\begin{table}[t]
  \centering
  \begin{minipage}[t]{0.45\linewidth}
    \centering
    \caption{Ablation on design choices. ``Adjust Magn.'' denotes whether the magnitude matrix is updated, ``Adjust Dir.'' denotes whether the direction matrix is updated, and ``Inject Orth.'' indicates whether the direction matrix is frozen and rotated via orthogonal matrices.}
    \setlength{\tabcolsep}{3pt}
    \resizebox{\linewidth}{!}{
      \begin{tabular}{ccc|c|c}
        \toprule
        Adjust Magn.? & Adjust Dir.? & Inject Orth.? & Mean$\downarrow$ & Gain$\uparrow$ \\
        \midrule
        \multicolumn{3}{c|}{Baseline} & 55.8 & 0.0 \\
        \midrule
        \checkmark & \ding{55} & \ding{55} & 46.0 & +9.8 \\
        \ding{55}  & \checkmark & \ding{55} & 51.3 & +4.5 \\
        \checkmark & \checkmark & \ding{55} & 52.2 & +3.6 \\
        \ding{55}  & \ding{55} & \checkmark & 44.1 & +11.7 \\
        \checkmark & \ding{55} & \checkmark & 42.2 & +13.6 \\
        \midrule
        \multicolumn{3}{c|}{LoRA~\cite{hu2022lora}} & 49.7 & +6.1 \\
        \multicolumn{3}{c|}{DoRA~\cite{liu2024dora}} & 48.1 & +7.7 \\
        \bottomrule
      \end{tabular}
    }
    \label{tab:component}
  \end{minipage}
  \hfill
  \begin{minipage}[t]{0.51\linewidth}
    \centering
    \caption{Computational analysis of different methods. ``\#Param.'' denotes the number of learnable parameters, while ``\#Extra Param.'' refers to additional parameters introduced during CTTA. ``\#FP'' and ``\#BP'' indicate the forward and backward propagation times, respectively. ``Time'' represents the relative computation time (normalized by TENT).}
    \setlength{\tabcolsep}{4pt}
    \resizebox{\linewidth}{!}{
      \begin{tabular}{ccccccc}
        \toprule
        Method & \#Param. & \#Extra Param. & \#FP & \#BP & Time & Err. Mean$\downarrow$ \\
        \midrule
        TENT~\cite{wang2020tent}  & 0.03M  &        & 1   & 1   & 1.0 & 51.0 \\
        CoTTA~\cite{song2023ecotta} & 86.57M &        & 11.7 & 1 & 3.6 & 54.8 \\
        VDP~\cite{gan2023decorate}  & 1800   & \checkmark & 2   & 1   & 1.5 & 50.0 \\
        EcoTTA~\cite{song2023ecotta} & 3.46M & \checkmark & 1   & 1   & 1.9 & 48.0 \\
        ViDA~\cite{liu2023vida}     & 7.13M & \checkmark & 11  & 1   & 2.8 & 43.4 \\
        \midrule
        Ours                        & 1.24M & \checkmark & 1   & 1   & 1.6 & 42.2 \\
        \bottomrule
      \end{tabular}
    }
    \label{tab:computation}
  \end{minipage}
  \vspace{-0.2cm}
\end{table}

\begin{figure}[t]
    \centering
    \includegraphics[width=1.0\linewidth]{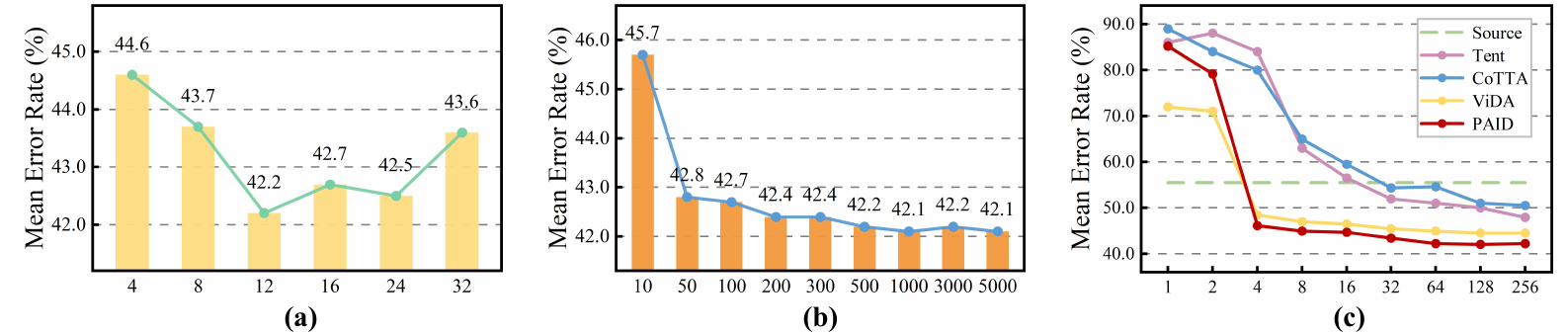}
    \vspace{-0.5cm}
    \caption{Ablation on (a) the coefficient of the injected orthogonal matrix, (b) the number of source domain samples, and (c) the test-time batch size}
    \label{batch_source}
    \vspace{-0.5cm}
\end{figure}

\subsection{Ablation Analysis}
\label{ablation_studies}

\noindent
{\bf Effect of Each Component.}
We conduct ablation studies to validate the effectiveness of key design choices in PAID, as shown in Tab.~\ref{tab:component}. A central finding is that preserving the pairwise angular structure during adaptation is essential for strong CTTA performance.
Directly adjusting the direction matrix, which allows each weight vector to rotate independently, disrupts the relative geometry and leads to performance degradation. 
In contrast, PAID freezes the original direction matrix and applies a learnable orthogonal transformation to rotate all vectors jointly. This enables adjustment of absolute angles while preserving their relative angular structure.
We also compare two general parameter-efficient tuning methods: LoRA~\cite{hu2022lora}, which adds low-rank matrices to linear layers, and DoRA~\cite{liu2024dora}, which decomposes linear weights into magnitude and direction and applies LoRA to the directional component. Both methods perform sub-optimally in our setting, suggesting that generalization in CTTA demands tailored, customized designs.

\noindent
{\bf Effect of Injected Matrix Coefficient.}
We perform an ablation study on the number of orthogonal matrices used in the Householder transformation chain (denoted as the coefficient $r$), as shown in Fig. \ref{batch_source} (a).
While increasing the number of matrices enhances the model's expressive capacity, we observe that CTTA performance does not improve monotonically with the number of matrices.
A possible reason is that the corrections required for domain perturbations do not demand overly complex transformations, and higher complexity may negatively affect model convergence in online adaptation. Our experiments reveal that the best performance is achieved when the number of matrices is set to 12, indicating an efficient trade-off between capacity and stability.

\noindent
{\bf Effect of Number of Source Examples.}
We investigate the sensitivity of PAID to the amount of source data by varying the number of images used to pre-compute source-domain statistics from 0 to 5,000. As shown in Fig.~\ref{batch_source} (b), our method achieves strong performance with as few as 500 source images. It is worth emphasizing that these statistics are computed once prior to CTTA and are not involved in the online adaptation process. Moreover, storing the computed statistics incurs negligible memory cost. These results demonstrate that PAID requires minimal source-domain information to function effectively, making it practical for real-world scenarios.

\noindent
{\bf Effect of Test Batch Size.}
To comprehensively evaluate the impact of test-time batch size, we compare various CTTA methods under batch sizes ranging from 1 to 256. As shown in Fig.~\ref{batch_source} (c), a consistent trend emerges across all methods: regardless of their objective functions, performance remains stable with sufficiently large batches but deteriorates as the batch size decreases. In the extreme case of single-sample adaptation, all methods suffer substantial performance drops.
In addition, our method maintains decent performance as long as the batch size exceeds 4, and consistently outperforms all comparison methods beyond this threshold.

\noindent
{\bf Analysis of Computation, Parameter, and Latency.}
We analyze the computational complexity of different methods in Tab.~\ref{tab:computation}, comparing the number of learnable parameters, the number of forward and backward passes, and relative runtime. While achieving the best overall performance, our method maintains a relatively small number of learnable parameters and avoids the repeated forward passes required by methods such as CoTTA, VDP, and ViDA. As a result, the increase in computational cost remains moderate, striking a good balance between efficiency and effectiveness.

\section{Conclusion}
\label{conclusion}
This work demonstrates that the pairwise angular structure of source pre-trained weights encodes the domain-invariant semantic prior, supported by statistical analysis, functional validation, visual evidence, and intuitive explanation. Leveraging this insight, we decompose weights into magnitude and direction, allowing magnitudes to adapt freely while constraining directional updates to global rotations via chained Householder transformations. This preserves the intrinsic angular structure during adaptation. Extensive experiments show that this structure-preserving strategy leads to consistently improved CTTA performance.

\bibliographystyle{plain}
\bibliography{ref}

\clearpage
\newpage
\setcounter{section}{0}
\begin{appendix}

\section{Additional Details of Three Motivation Experiments}
\label{motivation_experiment_detail}
In Experiment 1, we compute a weighted average of three statistics (mean, variance, and hyperspherical energy) across all linear layers, followed by min-max normalization across different cross-domain settings to produce the visualization in Fig. \ref{motivation1}. The TTA results correspond to our proposed method, PAID, applied under a non-continual setting where all linear layers are adapted. The fine-tune baseline refers to supervised fine-tuning of the pret-rained model on the CIFAR-100 dataset for 3 epochs, using warm-up and cosine annealing learning rate schedules. All linear layers and the classification head are learnable during this process.
In Experiment 2, the orthogonal rotation is implemented using the chained orthogonal matrices described later in the paper. All hyperparameter settings are aligned with those specified for ImageNet-C in our implementation details.
For the attention map visualization in Experiment 3, we use the attention-rollout codebase.\footnote{\url{https://github.com/BoCtrl-C/attention-rollout}}

\begin{figure}[htbp]
    \centering
    \includegraphics[width=1.0\linewidth]{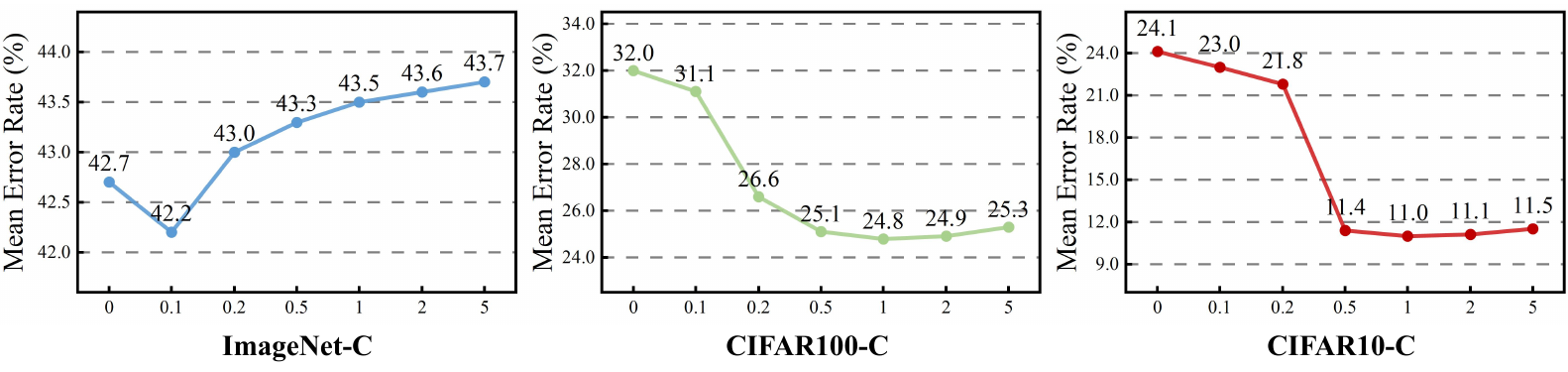}
    \caption{Ablation on loss balancing coefficient $\lambda$}
    \label{loss}
\end{figure}

\begin{table}[htbp]
  \centering
  \setlength{\tabcolsep}{3pt}
  \caption{Ablation on the selection of linear layer injection}
  \resizebox{\linewidth}{!}{
    \begin{tabular}{c|ccccccccccccccc|c}
    \toprule
    Ablation & Gaussian & Shot  & Impulse & Defocus & Glass & Motion & Zoom  & Snow  & Frost & Fog   & Bright & Contrast & Elastic & Pixel & JPEG  & Mean$\downarrow$ \\
    \midrule
    qv    & 51.1  & 44.2  & 44.5  & 60.0  & 58.3  & 50.3  & 50.8  & 40.8  & 41.7  & 49.4  & 25.3  & 59.0  & 43.4  & 37.1  & 35.8  & 46.1  \\
    qk    & 52.3  & 48.5  & 47.2  & 66.2  & 67.7  & 56.9  & 60.1  & 46.1  & 44.7  & 58.8  & 28.1  & 77.8  & 49.8  & 40.1  & 39.4  & 52.2  \\
    qkv   & 50.7  & 44.5  & 44.2  & 62.8  & 59.1  & 53.0  & 53.5  & 43.4  & 43.4  & 54.1  & 27.0  & 64.1  & 43.8  & 37.5  & 37.3  & 47.9  \\
    qkvo  & 48.8  & 43.0  & 43.2  & 58.5  & 56.8  & 51.6  & 50.0  & 41.0  & 42.2  & 48.5  & 26.7  & 57.8  & 44.0  & 38.6  & 37.5  & 45.9  \\
    qkvom & 48.8  & 43.7  & 44.4  & 49.4  & 49.6  & 47.3  & 44.2  & 37.5  & 39.4  & 42.1  & 25.2  & 50.0  & 39.3  & 35.5  & 36.5  & 42.2  \\
    \bottomrule
    \end{tabular}%
    }
  \label{tab:layer}%
\end{table}%

\begin{table}[htbp]
  \centering
  \setlength{\tabcolsep}{3pt}
  \caption{10-Round CTTA classification results on ImageNet-C, CIFAR100-C, and CIFAR10-C.}
  \resizebox{\linewidth}{!}{
    \begin{tabular}{c|c|ccccccccccccccc|c}
    \toprule
    Dataset & Setting & Gaussian & Shot  & Impulse & Defocus & Glass & Motion & Zoom  & Snow  & Frost & Fog   & Bright & Contrast & Elastic & Pixel & JPEG  & Mean$\downarrow$ \\
    \midrule
    \multirow{2}[1]{*}{ImageNet-C} & 1-Round & 48.8  & 43.7  & 44.4  & 49.4  & 49.6  & 47.3  & 44.2  & 37.5  & 39.4  & 42.1  & 25.2  & 50.0  & 39.3  & 35.5  & 36.5  & 42.2  \\
          & 10-Round & 47.0  & 43.8  & 43.5  & 48.6  & 50.6  & 46.2  & 42.8  & 39.5  & 38.8  & 40.5  & 25.5  & 47.2  & 39.2  & 34.8  & 36.4  & 41.6  \\
    \midrule
    \multirow{2}[0]{*}{CIFAR100-C} & 1-Round & 40.7  & 31.9  & 20.4  & 19.8  & 35.9  & 23.0  & 16.3  & 20.5  & 18.2  & 25.3  & 12.6  & 19.8  & 29.4  & 28.2  & 31.3  & 24.9  \\
          & 10-Round & 35.5  & 31.9  & 18.6  & 19.5  & 35.7  & 25.1  & 16.7  & 18.1  & 19.3  & 21.5  & 12.1  & 19.5  & 26.2  & 29.7  & 35.2  & 24.3  \\
    \midrule
    \multirow{2}[1]{*}{CIFAR10-C} & 1-Round & 22.9  & 11.8  & 9.9   & 9.1   & 16.7  & 10.8  & 7.4   & 7.4   & 6.6   & 11.4  & 4.5   & 9.3   & 12.8  & 9.4   & 14.5  & 11.0  \\
          & 10-Round & 15.9  & 13.5  & 9.0   & 9.3   & 15.3  & 9.1   & 8.9   & 7.0   & 6.7   & 10.2  & 4.2   & 8.2   & 15.8  & 9.6   & 12.0  & 10.3  \\
    \bottomrule
    \end{tabular}%
    }
  \label{tab:10_round}%
\end{table}%

\section{Additional Ablation Studies}
\label{additional_ablation_studies}

\noindent
{\bf Effect of Loss Balancing Coefficient.}
Unlike the orthogonal matrix coefficient $r$, which shows a consistent optimal value across the three classification benchmarks, the optimal value of the loss balancing coefficient $\lambda$ exhibits outliers.  Fig.~\ref{loss} show that ImageNet‑C achieves optimal accuracy with a much smaller $\lambda$ than CIFAR100‑C and CIFAR10‑C. 
ImageNet‑C keeps its native resolution of 224 $\times$ 224, and most corruptions simply shift global intensity, so aligning feature means removes most of the domain shift and only a small weight on variance alignment is needed. By contrast, CIFAR images are first enlarged from 32 $\times$ 32 to 384 $\times$ 384, which spreads each pixel and magnifies local artifacts, making second‑order statistics more important; therefore, a larger $\lambda$ that emphasizes variance alignment is required on the CIFAR variants.

\noindent
{\bf Effect of Linear Layer Injection.}
We examine where to place the orthogonal update in Tab.~\ref{tab:layer}. ViT-Base contains five types of linear layers: q, k, v, o (attention output), and m (MLP). We find that the best performance is achieved when all linear layers are updated (qkvom), indicating that distributing the orthogonal correction across both attention and feed-forward paths is essential. This suggests that corruption-induced degradation affects the model in a layer-wise and cumulative manner, and only full-layer adaptation can effectively counteract its impact.

\noindent
{\bf 10-Round Classification CTTA.}
To further evaluate the stability and effectiveness of our method, we conduct 10-round CTTA experiments on ImageNet-C, CIFAR100-C, and CIFAR10-C. Specifically, we cycle through the fifteen corruption domains ten times. As shown in Tab.~\ref{tab:10_round}, our method remains stable throughout and even achieves slight performance improvements as adaptation progresses. The values reported in the "10-round" row of Tab.~\ref{tab:10_round} represent the average results over the ten rounds.

\section{Limitations}
\label{limitation}
While PAID shows consistent gains under diverse corruptions, it is built upon a central hypothesis: the angular structure learned by source pre-trained models is potentially generalizable across diverse target domains. This assumption has been preliminarily validated through experiments, yet its applicability boundaries remain unclear, particularly under extreme conditions. Moreover, hyper-parameter tuning remains a common challenge in the CTTA community, underscoring the need for automated tuning strategies to enhance practical usability. Finally, extending the idea of angular structure preservation to tasks such as few-shot learning, and continual learning presents a promising direction, where maintaining the geometric stability of weights may likewise prove essential.

\section{Fine-grained CTTA Performance}
\label{fine-grained}
In this section, we extend the classification and segmentation results reported in our submission by presenting a detailed, fine-grained performance analysis. Specifically, we evaluate classification error rate and average mIoU score across distinct corruption types. To complement the summary results shown in Tab.~\ref{tab:imagenet-mean-gain} to Tab.~\ref{tab:acdc-mean-gain}, we provide additional detailed results in Tab.~\ref{tab:imagenet} to Tab.~\ref{tab:acdc}. These comprehensive evaluations demonstrate the robustness and effectiveness of our approach in various CTTA scenarios, including ImageNet-to-ImageNet-C, CIFAR-10-to-CIFAR-10-C, CIFAR-100-to-CIFAR-100-C, and Cityscapes-to-ACDC.

\begin{table}[t]
  \centering
  \setlength{\tabcolsep}{3pt} 
  \caption{Classification error rate (\%) for ImageNet-to-ImageNet-C, evaluated on ViT-Base with corruption severity level 5. \textbf{Bold} indicates the best performance.}
  \resizebox{\linewidth}{!}{
    \begin{tabular}{l|ccccccccccccccc|cc}
    \toprule
    \rotatebox[origin=c]{0}{Method} & \rotatebox[origin=c]{60}{Gaussian} & \rotatebox[origin=c]{60}{Shot} & \rotatebox[origin=c]{60}{Impulse} & \rotatebox[origin=c]{60}{Defocus} & \rotatebox[origin=c]{60}{Glass} & \rotatebox[origin=c]{60}{Motion} & \rotatebox[origin=c]{60}{Zoom} & \rotatebox[origin=c]{60}{Snow} & \rotatebox[origin=c]{60}{Frost} & \rotatebox[origin=c]{60}{Fog} & \rotatebox[origin=c]{60}{Bright} & \rotatebox[origin=c]{60}{Contrast} & \rotatebox[origin=c]{60}{Elastic} & \rotatebox[origin=c]{60}{Pixel} & \rotatebox[origin=c]{60}{JPEG}  & \rotatebox[origin=c]{0}{Mean$\downarrow$}  & \rotatebox[origin=c]{0}{Gain$\uparrow$} \\
    \midrule
    Source & 53.0  & 51.8  & 52.1  & 68.5  & 78.8  & 58.5  & 63.3  & 49.9  & 54.2  & 57.7  & 26.4  & 91.4  & 57.5  & 38.0  & 36.2  & 55.8  & 0.0  \\
    Pseudo~\cite{lee2013pseudo} & \textbf{45.2}  & \textbf{40.4}  & \textbf{41.6}  & 51.3  & 53.9  & 45.6  & 47.7  & 40.4  & 45.7  & 93.8  & 98.5  & 99.9  & 99.9  & 98.9  & 99.6  & 66.8  & -11.0  \\
    TENT~\cite{wang2020tent}  & 52.2  & 48.9  & 49.2  & 65.8  & 73.0  & 54.5  & 58.4  & 44.0  & 47.7  & 50.3  & 23.9  & 72.8  & 55.7  & 34.4  & 33.9  & 51.0  & +4.8  \\
    CoTTA~\cite{wang2022continual} & 52.9  & 51.6  & 51.4  & 68.3  & 78.1  & 57.1  & 62.0  & 48.2  & 52.7  & 55.3  & 25.9  & 90.0  & 56.4  & 36.4  & 35.2  & 54.8  & +1.0  \\
    VDP~\cite{gan2023decorate}   & 52.7  & 51.6  & 50.1  & 58.1  & 70.2  & 56.1  & 58.1  & 42.1  & 46.1  & 45.8  & 23.6  & 70.4  & 54.9  & 34.5  & 36.1  & 50.0  & +5.8  \\
    SAR~\cite{niu2023towards}   & 45.8  & 45.9  & 47.7  & 52.3  & 63.7  & 46.2  & 50.9  & 40.3  & 42.4  & \textbf{41.8}  & 24.4  & 53.4  & 53.6  & 38.4  & 36.6  & 45.6  & +10.2  \\
    RoTTA~\cite{yuan2023robust} & 51.5  & 50.3  & 51.7  & 60.4  & 58.7  & 52.6  & 54.8  & 47.2  & 43.5  & 42.8  & 25.9  & 49.1  & 48.8  & 46.3  & 39.7  & 48.2  & +7.6  \\
    EcoTTA~\cite{song2023ecotta} & 48.1  & 45.6  & 46.3  & 56.5  & 67.1  & 50.4  & 57.1  & 41.3  & 44.5  & 43.8  & \textbf{24.1}  & 71.6  & 54.8  & 34.1  & 34.8  & 48.0  & +7.8  \\
    ViDA~\cite{liu2023vida}  & 47.7  & 42.5  & 42.9  & 52.2  & 56.9  & 45.5  & 48.9  & 38.9  & 42.7  & 40.7  & 24.3  & 52.8  & 49.1  & 33.5  & 32.3  & 43.4  & +12.4  \\
    C-MAE~\cite{liu2024continual} & 46.3  & 41.9  & 42.5  & 51.4  & 54.9  & \textbf{43.3}  & \textbf{40.7}  & \textbf{34.2}  & \textbf{35.8}  & 64.3  & 23.4  & 60.3  & \textbf{37.5}  & \textbf{29.2}  & \textbf{31.4}  & 42.5  & +13.3  \\
    \midrule
    Ours  & 48.8  & 43.7  & 44.4  & \textbf{49.4}  & \textbf{49.6}  & 47.3  & 44.2  & 37.5  & 39.4  & 42.1  & 25.2  & \textbf{50.0}  & 39.3  & 35.5  & 36.5  & \textbf{42.2}  & \textbf{+13.6}  \\
    \bottomrule
    \end{tabular}%
    }
  \label{tab:imagenet}%
\end{table}%

\begin{table}[t]
  \centering
  \setlength{\tabcolsep}{3pt} 
  \caption{Classification error rate (\%) for CIFAR100-to-CIFAR100-C, evaluated on ViT-Base with corruption severity level 5. \textbf{Bold} indicates the best performance.}
  \resizebox{\linewidth}{!}{
    \begin{tabular}{l|ccccccccccccccc|cc}
    \toprule
    \rotatebox[origin=c]{0}{Method} & \rotatebox[origin=c]{60}{Gaussian} & \rotatebox[origin=c]{60}{Shot} & \rotatebox[origin=c]{60}{Impulse} & \rotatebox[origin=c]{60}{Defocus} & \rotatebox[origin=c]{60}{Glass} & \rotatebox[origin=c]{60}{Motion} & \rotatebox[origin=c]{60}{Zoom} & \rotatebox[origin=c]{60}{Snow} & \rotatebox[origin=c]{60}{Frost} & \rotatebox[origin=c]{60}{Fog} & \rotatebox[origin=c]{60}{Bright} & \rotatebox[origin=c]{60}{Contrast} & \rotatebox[origin=c]{60}{Elastic} & \rotatebox[origin=c]{60}{Pixel} & \rotatebox[origin=c]{60}{JPEG}  & \rotatebox[origin=c]{0}{Mean$\downarrow$}  & \rotatebox[origin=c]{0}{Gain$\uparrow$} \\
    \midrule
    Source & 55.0  & 51.5  & 26.9  & 24.0  & 60.5  & 29.0  & 21.4  & 21.1  & 25.0  & 35.2  & 11.8  & 34.8  & 43.2  & 56.0  & 35.9  & 35.4  & 0.0  \\
    Pseudo~\cite{lee2013pseudo} & 53.8  & 48.9  & 25.4  & 23.0  & 58.7  & 27.3  & 19.6  & 20.6  & 23.4  & 31.3  & 11.8  & 28.4  & 39.6  & 52.3  & 33.9  & 33.2  & +2.2  \\
    TENT~\cite{wang2020tent}  & 53.0  & 47.0  & 24.6  & 22.3  & 58.5  & 26.5  & 19.0  & 21.0  & 23.0  & 30.1  & 11.8  & 25.2  & 39.0  & 47.1  & 33.3  & 32.1  & +3.3  \\
    CoTTA~\cite{wang2022continual} & 55.0  & 51.3  & 25.8  & 24.1  & 59.2  & 28.9  & 21.4  & 21.0  & 24.7  & 34.9  & 11.7  & 31.7  & 40.4  & 55.7  & 35.6  & 34.8  & +0.6  \\
    VDP~\cite{gan2023decorate}   & 54.8  & 51.2  & 25.6  & 24.2  & 59.1  & 28.8  & 21.2  & 20.5  & 23.3  & 33.8  & \textbf{7.5}   & \textbf{11.7}  & 32.0  & 51.7  & 35.2  & 32.0  & +3.4  \\
    ViDA~\cite{liu2023vida}  & 50.1  & 40.7  & 22.0  & 21.2  & 45.2  & \textbf{21.6}  & 16.5  & \textbf{17.9}  & \textbf{16.6}  & 25.6  & 11.5  & 29.0  & 29.6  & 34.7  & 27.1  & 27.3  & +8.1  \\
    C-MAE~\cite{liu2024continual} & 48.6  & \textbf{30.7}  & \textbf{18.5}  & 21.3  & 38.4  & 22.2  & 17.5  & 19.3  & 18.0  & \textbf{24.8}  & 13.1  & 27.8  & 31.4  & 35.5  & \textbf{29.5}  & 26.4  & +9.0  \\
    \midrule
    Ours  & \textbf{40.7}  & 31.9  & 20.4  & \textbf{19.8}  & \textbf{35.9}  & 23.0  & \textbf{16.3}  & 20.5  & 18.2  & 25.3  & 12.6  & 19.8  & \textbf{29.4}  & \textbf{28.2}  & 31.3  & \textbf{24.9}  & \textbf{+10.5}  \\
    \bottomrule
    \end{tabular}%
    }
  \label{tab:cifar100}%
\end{table}%

\begin{table}[t]
  \centering
  \setlength{\tabcolsep}{3pt} 
  \caption{Classification error rate (\%) for CIFAR10-to-CIFAR10-C, evaluated on ViT-Base with corruption severity level 5. \textbf{Bold} indicates the best performance.}
  \resizebox{\linewidth}{!}{
    \begin{tabular}{l|ccccccccccccccc|cc}
    \toprule
    \rotatebox[origin=c]{0}{Method} & \rotatebox[origin=c]{60}{Gaussian} & \rotatebox[origin=c]{60}{Shot} & \rotatebox[origin=c]{60}{Impulse} & \rotatebox[origin=c]{60}{Defocus} & \rotatebox[origin=c]{60}{Glass} & \rotatebox[origin=c]{60}{Motion} & \rotatebox[origin=c]{60}{Zoom} & \rotatebox[origin=c]{60}{Snow} & \rotatebox[origin=c]{60}{Frost} & \rotatebox[origin=c]{60}{Fog} & \rotatebox[origin=c]{60}{Bright} & \rotatebox[origin=c]{60}{Contrast} & \rotatebox[origin=c]{60}{Elastic} & \rotatebox[origin=c]{60}{Pixel} & \rotatebox[origin=c]{60}{JPEG}  & \rotatebox[origin=c]{0}{Mean$\downarrow$}  & \rotatebox[origin=c]{0}{Gain$\uparrow$} \\
    \midrule
    Source & 60.1  & 53.2  & 38.3  & 19.9  & 35.5  & 22.6  & 18.6  & 12.1  & 12.7  & 22.8  & 5.3   & 49.7  & 23.6  & 24.7  & 23.1  & 28.1  & 0.0  \\
    Pseudo~\cite{lee2013pseudo} & 59.8  & 52.5  & 37.2  & 19.8  & 35.2  & 21.8  & 17.6  & 11.6  & 12.3  & 20.7  & 5.0   & 41.7  & 21.5  & 25.2  & 22.1  & 26.9  & +1.2  \\
    TENT~\cite{wang2020tent}  & 57.7  & 56.3  & 29.4  & 16.2  & 35.3  & 16.2  & 12.4  & 11.0  & 11.6  & 14.9  & 4.7   & 22.5  & 15.9  & 29.1  & 19.5  & 23.5  & +4.6  \\
    CoTTA~\cite{wang2022continual} & 58.7  & 51.3  & 33.0  & 20.1  & 34.8  & 20.0  & 15.2  & 11.1  & 11.3  & 18.5  & 4.0   & 34.7  & 18.8  & 19.0  & 17.9  & 24.6  & +3.5  \\
    VDP~\cite{gan2023decorate}   & 57.5  & 49.5  & 31.7  & 21.3  & 35.1  & 19.6  & 15.1  & 10.8  & 10.3  & 18.1  & 4.0   & 27.5  & 18.4  & 22.5  & 19.9  & 24.1  & +4.0  \\
    ViDA~\cite{liu2023vida}  & 52.9  & 47.9  & 19.4  & 11.4  & 31.3  & 13.3  & 7.6   & 7.6   & 9.9   & 12.5  & \textbf{3.8}   & 26.3  & 14.4  & 33.9  & 18.2  & 20.7  & +7.4  \\
    C-MAE~\cite{liu2024continual} & 30.6  & 18.9  & 11.5  & 10.4  & 22.5  & 13.9  & 9.8   & 6.6   & \textbf{6.5}   & \textbf{8.8}   & 4.0   & \textbf{8.5}   & \textbf{12.7}  & \textbf{9.2}   & \textbf{14.4}  & 12.6  & +15.5  \\
    \midrule
    Ours  & \textbf{22.9}  & \textbf{11.8}  & \textbf{9.9}   & \textbf{9.1}   & \textbf{16.7}  & \textbf{10.8}  & \textbf{7.4}   & 7.4   & 6.6   & 11.4  & 4.5   & 9.3   & 12.8  & 9.4   & 14.5  & \textbf{11.0}  & \textbf{+17.1}  \\
    \bottomrule
    \end{tabular}%
    }
  \label{tab:cifar10}%
\end{table}%

\begin{table}[t]
  \centering
  \caption{Average mIoU score (\%) for Cityscapes-to-ACDC, evaluated on Segformer-B5. The same target domains are repeated three rounds. \textbf{Bold} indicates the best performance.}
  \setlength{\tabcolsep}{2.5pt} 
  \resizebox{\linewidth}{!}{
    \begin{tabular}{l|ccccc|ccccc|ccccc|cc}
    \toprule
    \multirow{2}[2]{*}{Method} & \multicolumn{5}{c|}{Round 1}          & \multicolumn{5}{c|}{Round 2}          & \multicolumn{5}{c|}{Round 3}          & \multirow{2}[2]{*}{Mean$\uparrow$} & \multirow{2}[2]{*}{Gain$\uparrow$} \\
\cmidrule{2-16}          & Fog   & Night & Rain  & Snow  & Mean$\uparrow$  & Fog   & Night & Rain  & Snow  & Mean$\uparrow$  & Fog   & Night & Rain  & Snow  & Mean$\uparrow$  &       &  \\
    \midrule
    Source & 69.1  & 40.3  & 59.7  & 57.8  & 56.7  & 69.1  & 40.3  & 59.7  & 57.8  & 56.7  & 69.1  & 40.3  & 59.7  & 57.8  & 56.7  & 56.7  & 0.0  \\
    TENT~\cite{wang2020tent}  & 69.0  & 40.2  & 60.1  & 57.3  & 56.7  & 68.3  & 39.0  & 60.1  & 56.3  & 55.9  & 67.5  & 37.8  & 59.6  & 55.0  & 55.0  & 55.7  & -1.0  \\
    CoTTA~\cite{wang2022continual} & 70.9  & 41.2  & 62.4  & 59.7  & 58.6  & 70.9  & 41.1  & 62.6  & 59.7  & 58.6  & 70.9  & 41.0  & 62.7  & 59.7  & 58.6  & 58.6  & +1.9  \\
    DePT~\cite{gao2022visual}  & 71.0  & 40.8  & 58.2  & 56.8  & 56.5  & 68.2  & 40.0  & 55.4  & 53.7  & 54.3  & 66.4  & 38.0  & 47.3  & 47.2  & 49.7  & 53.4  & -3.3  \\
    VDP~\cite{gan2023decorate}   & 70.5  & 41.1  & 62.1  & 59.5  & 58.3  & 70.4  & 41.1  & 62.2  & 59.4  & 58.2  & \textbf{70.4}  & 41.0  & 62.2  & 59.4  & 58.2  & 58.2  & +1.5  \\
    SAR~\cite{niu2023towards}   & 69.0  & 40.2  & 60.1  & 57.3  & 56.7  & 69.0  & 40.3  & 60.0  & \textbf{67.8}  & 59.3  & 67.5  & 37.8  & 59.6  & 55.0  & 55.0  & 57.0  & +0.3  \\
    EcoTTA~\cite{song2023ecotta} & 68.5  & 35.8  & 62.1  & 57.4  & 56.0  & 68.3  & 35.5  & 62.3  & 57.4  & 55.9  & 68.1  & 35.3  & 62.3  & 57.3  & 55.8  & 55.8  & -0.9  \\
    ViDA~\cite{liu2023vida}  & 71.6  & 43.2  & 66.0  & \textbf{63.4}  & 61.1  & \textbf{73.2}  & 44.5  & 67.0  & 63.9  & 62.2  & 73.2  & 44.6  & 67.2  & \textbf{64.2}  & 62.3  & 61.9  & +5.2  \\
    C-MAE~\cite{liu2024continual} & \textbf{71.9}  & 44.6  & 67.4  & 63.2  & \textbf{61.8}  & 71.7  & 44.9  & 66.5  & 63.1  & 61.6  & 72.3  & 45.4  & 67.1  & 63.1  & 62.0  & 61.8  & +5.1  \\
    \midrule
    Ours  & 69.6  & \textbf{45.5}  & \textbf{68.0}  & 60.7  & 61.0  & 72.3  & \textbf{45.2}  & \textbf{66.8}  & 62.5  & \textbf{61.7}  & 72.6  & \textbf{46.9}  & \textbf{68.4}  & 63.8  & \textbf{62.9}  & \textbf{62.2}  & \textbf{+5.5}  \\
    \bottomrule
    \end{tabular}%
    }
  \label{tab:acdc}%
\end{table}%

\end{appendix}

\end{document}